\documentclass[sigconf]{acmart}

\AtBeginDocument{
	}

\acmConference[KDD '25]{the 31th ACM SIGKDD Conference on Knowledge Discovery and Data Mining}{August 25--29,
	2025}{Toronto, Canada}
\acmISBN{978-1-4503-XXXX-X/18/06}

\copyrightyear{2025}
\acmYear{2025}
\setcopyright{acmlicensed}\acmConference[KDD '25]{Proceedings of the 31st ACM SIGKDD Conference on Knowledge Discovery and Data Mining V.1}{August 3--7, 2025}{Toronto, ON, Canada}
\acmBooktitle{Proceedings of the 31st ACM SIGKDD Conference on Knowledge Discovery and Data Mining V.1 (KDD '25), August 3--7, 2025, Toronto, ON, Canada}
\acmDOI{10.1145/3690624.3709185}
\acmISBN{979-8-4007-1245-6/25/08}

\usepackage{subfigure}
\usepackage{algorithmic}
\usepackage{algorithm}
\usepackage{lipsum}
\usepackage{multicol}
\usepackage{multirow}

\usepackage{amsmath,amsfonts,bm}

\newcommand{\newterm}[1]{{\bf #1}}

\def\eqref#1{equation~\ref{#1}}

\def\1{\bm{1}}

\def\vtheta{{\bm{\theta}}}
\def\va{{\bm{a}}}

\def\vh{{\bm{h}}}

\def\vp{{\bm{p}}}

\def\vx{{\bm{x}}}
\def\vy{{\bm{y}}}

\def\mH{{\bm{H}}}

\def\mM{{\bm{M}}}

\def\mP{{\bm{P}}}

\def\mW{{\bm{W}}}
\def\mX{{\bm{X}}}

\DeclareMathAlphabet{\mathsfit}{\encodingdefault}{\sfdefault}{m}{sl}
\SetMathAlphabet{\mathsfit}{bold}{\encodingdefault}{\sfdefault}{bx}{n}

\def\gE{{\mathcal{E}}}

\def\gG{{\mathcal{G}}}

\def\gL{{\mathcal{L}}}

\def\gV{{\mathcal{V}}}

\def\sF{{\mathbb{F}}}

\def\sP{{\mathbb{P}}}

\def\sU{{\mathbb{U}}}
\def\sV{{\mathbb{V}}}

\newcommand{\R}{\mathbb{R}}

\newcommand{\gnn}{TPGNN}
\newcommand{\ours}{\texttt{TransPlace}}
\makeatletter
\newenvironment{breakablealgorithm}
  {
   \begin{center}
     \refstepcounter{algorithm}
     \hrule height.8pt depth0pt \kern2pt
     \renewcommand{\caption}[2][\relax]{
       {\raggedright\textbf{\fname@algorithm~\thealgorithm} ##2\par}%
       \ifx\relax##1\relax 
         \addcontentsline{loa}{algorithm}{\protect\numberline{\thealgorithm}##2}%
       \else 
         \addcontentsline{loa}{algorithm}{\protect\numberline{\thealgorithm}##1}%
       \fi
       \kern2pt\hrule\kern2pt
     }
  }{
     \kern2pt\hrule\relax
   \end{center}
  }
\makeatother

\usepackage[nameinlink,capitalise]{cleveref}
\crefname{section}{§}{§§}
\Crefname{section}{§}{§§}
\crefname{lemma}{lemma}{lemmas}
\Crefname{lemma}{Lemma}{Lemmas}
\crefname{thm}{theorem}{theorems}
\Crefname{thm}{Theorem}{Theorems}

\begin{document}

\title{TransPlace: Transferable Circuit Global Placement via Graph Neural Network}

\author{Yunbo Hou}
\affiliation{
\department{School of Software and Microelectronics}
  \institution{Peking University}
  \city{Beijing}
  \country{China}
  }
\email{yunboh@stu.pku.edu.cn}

\author{Haoran Ye}
\affiliation{
\department[0]{National Key Laboratory of General Artificial Intelligence}
  \department[1]{School of Intelligence Science and Technology}
  \institution{Peking University}
  \city{Beijing}
  \country{China}
}
\email{hrye@stu.pku.edu.cn}

\author{Shuwen Yang}
\orcid{0009-0008-1358-9594}
\affiliation{
  \institution{DP Technology}
  \city{Beijing}
  \country{China}
}
\email{yangsw@dp.tech}

\author{Yingxue Zhang}
\affiliation{
 \institution{Huawei Noah’s Ark Lab}
 \city{Markham}
 \country{Canada}
 }
\email{yingxue.zhang@huawei.com}

\author{Siyuan Xu}
\affiliation{
  \institution{Huawei Noah’s Ark Lab}
  \city{Shenzhen}
  \country{China}
  }
\email{xusiyuan520@huawei.com}

\author{Guojie Song}
\authornote{Corresponding author.}
\affiliation{
\department[0]{National Key Laboratory of General Artificial Intelligence}
  \department[1]{School of Intelligence Science and Technology}
  \institution{Peking University}
  \city{Beijing}
  \country{China}
  }
\email{gjsong@pku.edu.cn}

\renewcommand{\shortauthors}{Trovato et al.}

\begin{abstract}

Global placement, a critical step in designing the physical layout of computer chips, is essential to optimize chip performance.
Prior global placement methods optimize each circuit design individually from scratch. Their neglect of transferable knowledge limits solution efficiency and chip performance as circuit complexity drastically increases.
This study presents TransPlace, a global placement framework that learns to place millions of mixed-size cells in continuous space.
TransPlace introduces i) Netlist Graph to efficiently model netlist topology, ii) Cell-flow and relative position encoding to learn SE(2)-invariant representation, iii) a tailored graph neural network architecture for informed parameterization of placement knowledge, and iv) a two-stage strategy for coarse-to-fine placement.
Compared to state-of-the-art placement methods, TransPlace—trained on a few high-quality placements—can place unseen circuits with 1.2x speedup while reducing congestion by 30\%, timing by 9\%, and wirelength by 5\%.

\end{abstract}


\maketitle

\section{Introduction}\label{sec:introduction}

The development of integrated circuits (ICs) has propelled technological advancement from single chips to complex computing systems. Placement, a crucial stage in the design flow of Integrated Circuits (ICs), arranges electronic components within circuit layouts. 
As placement constructs the fundamental geometry from a topological netlist, it drives later design steps and guides prior stages, serving as a critical determinant of final chip performance.

The placement problem can be described as follows. Consider a circuit design represented by a hypergraph $H=(V,E)$, where $V$ represents the set of cells (electronic units), and $E$ represents the set of nets (hyperedges) between these cells. Placement determines cell positions $\vx$ and $\vy$ to minimize wirelength while avoiding cell overlap. 
This problem typically penalizes the density and mathematically formulates as \cite{eplace,ntuplace3}
\begin{equation}
    \min\limits_{\vx, \vy} \quad \sum_{e\in E}WL(e;\vx,\vy) + \lambda D(\vx,\vy).
\end{equation}
Here, $WL(\cdot)$ is the wirelength cost function that evaluates the wirelength of any given net instance $e$, $D(\cdot, \cdot)$ is the density function to spread cells out in the layout, and $\lambda$ is a weighting coefficient.

Despite decades of development, existing global placers tackle each placement instance from scratch using heuristics or expertly crafted algorithms \cite{dreamplace, NTUPlace4h, eplace, hidap, timber, mdptree, min-cut}. {These optimization-based and non-learnable global placers cannot leverage existing placement experience, thus limiting solution efficiency and chip performance.} On the other hand, recent advances in transferable placers \cite{googleplace,chipformer} focus on the floorplanning stage, employing reinforcement learning (RL) to train placement policies. 
However, these methods do not effectively scale to placements involving millions of mixed-size cells, and they have not been successfully applied to placement problems in continuous space.

This paper presents \textbf{\ours{}}, the first learning-based approach for global placement involving millions of mixed-size cells in the continuous space.
We aim to leverage transferable placement knowledge to enhance both solution efficiency and placement quality. Learning to inductively place millions of cells presents significant challenges: (1) modeling large-scale and complex circuits into structured graphs; (2) extracting informed supervised signals from existing placements; (3) learning transferable knowledge from diverse heterogeneous circuit placements; and (4) adapting to circuit-specific constraints.

We address the above challenges through a series of techniques:
\begin{itemize}

    \item We introduce \textbf{Netlist Graph} to represent the topology of a netlist. In this graph, cells and nets are represented as two distinct types of nodes, with pins serving as the connections between these nodes. The original circuit is hierarchically partitioned and coarsened, under optimal time complexity, which allows for an efficient and scalable encoding for circuits with millions of cells.
    
    \item We propose \textbf{Cell-flow}, a directed acyclic graph (DAG) in which each edge represents the relative positional relationship between cells. We extract cell-flows from well-placed circuits as demonstration data and train our model to learn SE(2)-invariant representations.

    \item We design \textbf{T}ransferable \textbf{P}lacement \textbf{G}raph \textbf{N}eural \textbf{N}etwork (\textbf{\gnn{}}) to facilitate heterogeneous message passing within and across netlist graphs and cell-flows. \gnn{} effectively parameterizes transferable placement knowledge by preserving full circuit topology and incorporating cell-flow inductive biases.

    \item We couple inductive placement with optimization-based \textbf{Circuit-adaptive Fine-tuning} that adapts to unique characteristics of unseen circuits, such as their terminal positions and core placement area. This two-stage strategy improves placement quality by integrating efficient inductive placement and fine-grained optimization-based placement.
\end{itemize}

We summarize our contributions as follows: (1) We introduce \ours{}, the first learning-based framework for large-scale global placement. (2) We propose a series of techniques—Netlist Graph modeling, Cell-flow, \gnn{}, and a two-stage strategy—to enable and enhance large-scale inductive placement. (3) We comprehensively evaluate \ours{} across four standard benchmarks, demonstrating that \ours{} can surpass state-of-the-art global placers with a 1.2x speedup, 30\% less congestion, 9\% better timing, and 5\% shorter wirelength.

\begin{figure*}
    \centering
    \includegraphics{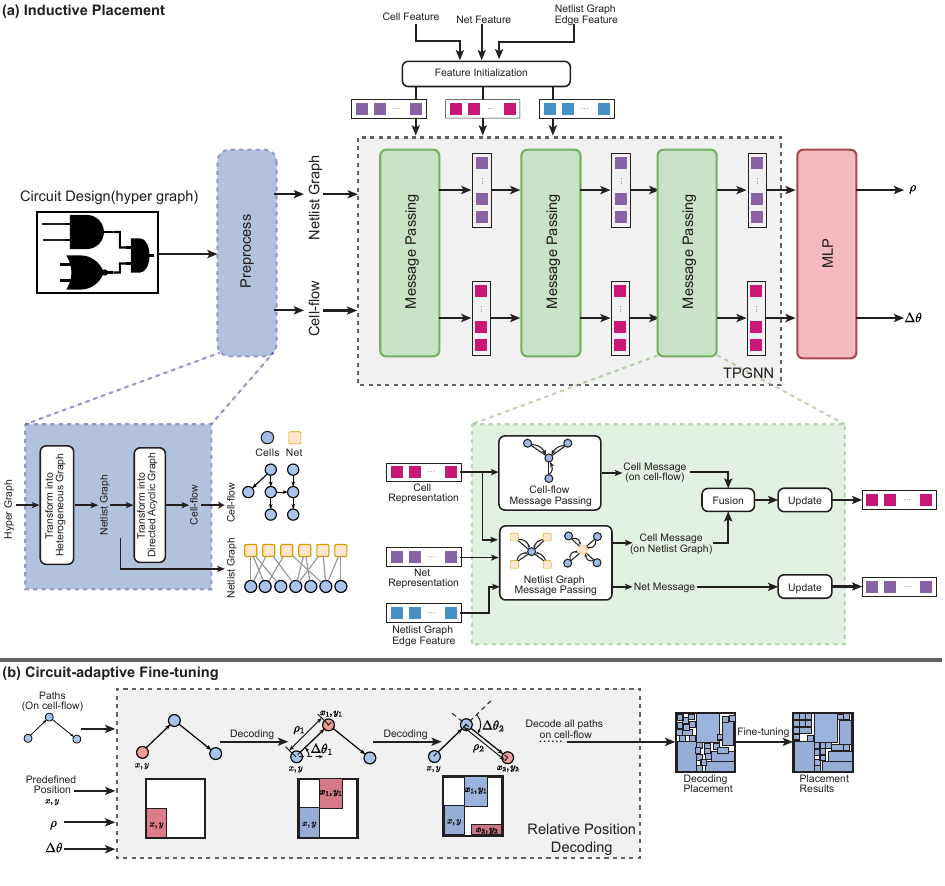}
    \caption{A schematic illustration of \ours{}. \ours{} contains two stages: Inductive Placement and Circuit-adaptive Fine-tuning. (a) Inductive Placement efficiently generates relative cell positions in one shot. It constructs the Cell-flow and Netlist Graph, applies message-passing on the two graphs to obtain cell and net representations, fuses and updates the hidden representations, and reads out relative position \(\rho, \Delta\theta\). (b) Circuit-adaptive Fine-tuning decodes the relative positions within cell-flow into absolute positions and performs iterative gradient-based optimization.}
\Description{}
\label{fig:method}

\end{figure*}

\begin{figure*}[htbp]
    \centering
    \includegraphics[width=0.99\linewidth]{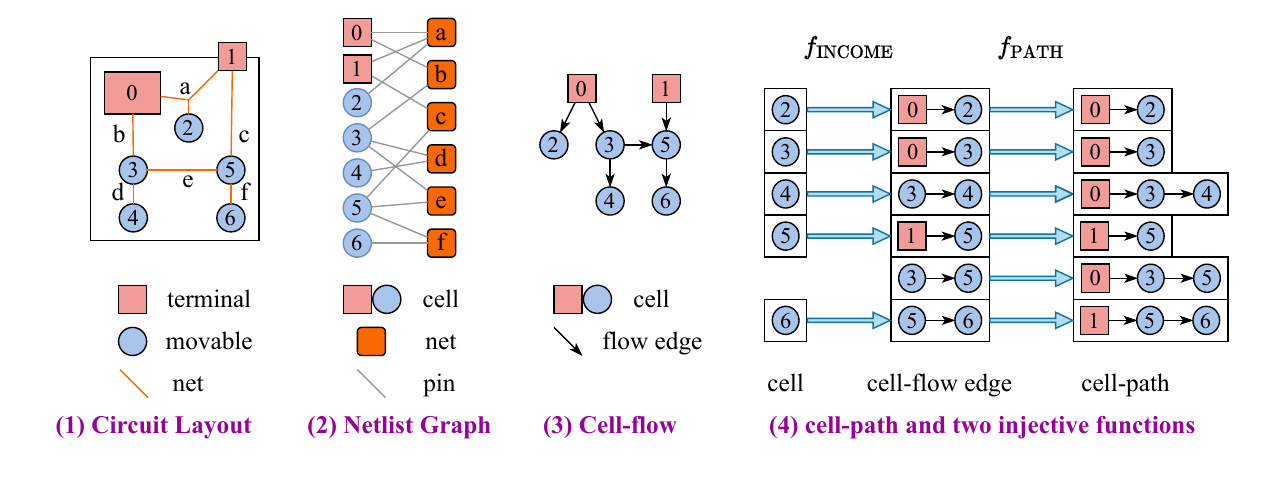}
	\caption{Overview of Netlist Graph, Cell-flow, cell-path, $f_{\rm INCOME}$, and $f_{\rm PATH}$.}
 \Description{}
    \label{fig:cell-flow}
\end{figure*}

\section{\ours{}}
\label{sec:method}

\ours{} is schematically illustrated in \cref{fig:method}, consisting of two stages: inductive placement and circuit-adaptive fine-tuning. For inductive placement, we introduce Netlist graph (\cref{sec:hier}) and cell-flow (\cref{sec:cell-flow}) for efficient and topology-aware circuit modeling, SE(2)-invariant encoding and decoding that converts between relative and absolute cell positions (\cref{sec:enc_dec}), and \gnn{} to learn transferable placement knowledge (\cref{sec:gnn}). Following inductive placement, we fine-tune the placement to adapt to circuit-specific characteristics and constraints (\cref{sec:fine-tuning}).


\subsection{Netlist Graph}
\label{sec:hier}
The circuit design can be represented as a hypergraph $H=(\gV, \gE)$, where $\gV$ denotes the set of cells (electronic units) and $\gE$ is the set of nets (hyperedges). $H$ stores the circuit topology produced in the preceding logic synthesis stage. As exemplified in \cref{fig:cell-flow}(1), there are two types of electronic cells $\sV$:

\begin{enumerate}
    \item \newterm{Terminals} $\sV_T$ are big chunks of cells with fixed positions.
    \item \newterm{Movable cells} $\sV_M$ are relatively smaller cells, and their positions are the decision variables of global placement.
\end{enumerate}

The Netlist Graph (\cref{fig:cell-flow}(2)), which serves as an input for inductive placement, is designed to fully retain the topological information. We formally define it as follows:

\begin{definition}[Netlist Graph]
    $\gG=\{\sV,\sU,\sP,\mX_\sV,\mX_\sU,\mX_\sP,\mP_T\}$, where $\sV=\sV_T\cup\sV_M$ are electronic cells, nets $\sU$ are the hyper-edges connecting the cells, and pins $\sP\subseteq\sV\times\sU$ represent the interactions among cells and nets. $\mX$s are their feature matrices. $\mP_T\in\R^{|\sV_T|\times 2}$ stores the horizontal/vertical positions of terminals $\sV_T$.
    \label{def:netlist}
\end{definition}

However, for some VLSIs with millions of cells and nets, directly processing the whole netlist graph exceeds memory limits and prevents effective learning. Therefore, we transform the original netlist graph $\gG$ to \newterm{Hierarchical Netlist Graph} $\widetilde\gG=f_{\rm HIER}(\gG)=(\gG_R,\{\gG_1,\cdots,$ $\gG_n\})$, where $\{\gG_i|i=1,\cdots,n\}$ are sub-netlist graphs of $\gG$ (named \newterm{branch graphs}) and $\gG_R$ is a transformation of $\gG$ coarsened with the branch graphs (named \newterm{root graph}). Specifically, we first produce a partition result with KaHyPar \cite{kahypar}:
\begin{equation}
\begin{aligned}
    &{\rm KaHyPar}(\gG)=\{\sV_i\subset\sV_M|i=1,\cdots,n\},\\
    &\text{where}\ \forall i\ne j,\sV_i\cap\sV_j=\emptyset .
\end{aligned}
\end{equation}
Note that the cells within each partition set are fully connected via nets. Otherwise, we further partition the branches. For every $\sV_i\subset\sV_M$, we extract \newterm{Sub-netlist Graph} and coarsen the original netlist graph; more algorithmic details are presented in \cref{sec:app-sub-netlist}. The training and inference are performed on individual graphs in $\widetilde\gG$, and the placement results of original $\gG$ are reconstructed from $\widetilde\gG$. In \cref{sec:opt-hier},
we demonstrate that the time complexity of \cref{alg:hier} is $O(|\sV|+|\sU|+|\sP|)$, which is optimal.

\subsection{Cell-flow}
\label{sec:cell-flow}
Learning (from) relative positional relationships instead of absolute spatial positions improves generalization across tasks \cite{satorras2021n, kim2023devformer, ye2023deepaco, du2022se, physchem}. However, the complexity of preserving all relative positions in global placement is \(\sum_e degree_e\), where \(e\) denotes the hyperedges. For circuits with millions of nodes and hyperedges, such naive modeling would cause infeasible computational demands. In answer to this, we present cell-flow to efficiently model relative positional relationships in circuit designs.

\newterm{Cell-flow} (\cref{fig:cell-flow}(3)) is a set of directed acyclic edges. It represents the relative positions of cells in a circuit for encoding (the demonstration data) and decoding (the inductive solution). We initialize a cell-flow from a netlist graph with a breadth-first search that starts from fixed cells and travels through nets. Meanwhile, we record the incoming flow edge for each cell using $f_{\rm INCOME}$, and the net containing the flow edge using $f_{\rm NET}$. This process is elaborated in \cref{alg:cell-flow}. Note that every connectivity branch contains at least one fixed cell (e.g. port), so cell-flow involves all cells. This is also guaranteed for sub-netlist graphs in \cref{sec:hier} because they are fully connected and contain a pseudo-terminal.

\begin{definition}[Cell-flow]
    A set of directed edges $\sF\subseteq\sV\times\sV$, where $\sV$ are electronic cells in netlist graph $\gG$. It is guaranteed that no loop will be found in $\sF$. 
    \label{def:cell-flow}
\end{definition}

As demonstrated in \cref{sec:opt-flow}, the time-consumption of generating cell-flow edges is $O(|\sV|+|\sU|+|\sP|)$, and the cell-flow size is $|\sF|=O(|\sP|)$. So this procedure is optimal in terms of time and space complexity.

\subsection{SE(2)-invariant Encoding and Decoding}
\label{sec:enc_dec}

Based on the relative position relationship within cell-flow, we introduce an SE(2)-invariant encoding/decoding algorithm to convert between cell absolute/relative positions. This SE(2) invariance denotes that the representations remain unchanged under rotations and translations in the 2D plane, which allows our model to learn informed representation by preserving geometric consistency.

For the demonstration data (labels), we encode the absolute cell positions $\mP$ into relative positions $\vec \mP$ for training. This encoding process follows the cell-flow $\sF$ within the placement and computes relative positions only for the flow edges $(v_i, v_j) \in \sF$:
\begin{equation}
    \vec\vp_{i,j}=\vp_j-\vp_i,\qquad (v_i,v_j)\in\sF,
    \label{eq:enc-p}
\end{equation}
where $\vp_i$ is the absolute position of cell $v_i$. The vector $\vec\vp_{i,j}$ of flow edge $(v_i,v_j)$ is later transformed into diameter $\rho_{i,j}$ and angle $\theta_{i,j}$ in polar coordinate, and a corresponding deflection $\Delta\theta_{i,j}$ is calculated by subtracting the polar angle of $v_i$'s cell-incoming-edge:
\begin{equation}
    \rho_{i,j}=|\vec\vp_{i,j}|_2\qquad\theta_{i,j} = \arctan\vec\vp_{i,j},
    \label{eq:enc-rho-theta}
\end{equation}
\begin{equation}
    \Delta\theta_{i,j}=\theta_{i,j}-\theta_{k,i},
    \qquad\text{where}\ (v_k,v_i)=f_{\rm INCOME}(v_i).
    \label{eq:enc-delta-theta}
\end{equation}
Overall, the distance $\rho$ and the deflection $\Delta\theta$ are utilized as the SE(2)-invariant encoding of the cell positions $\mP$: 
\begin{equation}
    (\bm{\rho},\Delta\vtheta)=f_{\rm ENCODE}(\mP_M;\gG).
    \label{eq:encode}
\end{equation}

On the other hand, we need to convert the inductively generated relative positions into absolute ones. In this decoding process, given an input netlist graph $\gG$ and a relative position encoding $(\bm{\rho}, \Delta\vtheta)$, we determine the absolute positions $\mP_M$ of the movable cells by extracting the cell-paths (see \cref{fig:cell-flow}(4)) from $\gG$ as described in \cref{sec:cell-flow}. For a cell-path $(v_0, v_1, \cdots, v_{t-1}, v_t)_P$, which starts from a fixed cell $v_0$ with a fixed position $\vp_0$, we can calculate the position vector $\vec\vp_{t-1,t}$ of the ending cell-flow edge $(v_{t-1},v_t)$:
\begin{equation}
    \rho_0=|\vp_0|_2\qquad\theta_0=\arctan\vp_0\qquad
    \theta_{t-1,t}=\theta_0+\sum_{i=1,\cdots,t}\Delta\theta_{i-1,i}
    \label{eq:dec-1}
\end{equation}
\begin{equation}
    \vec\vp_{t-1,t}=[
        \rho_{t-1,t}\cos\theta_{t-1,t},
        \rho_{t-1,t}\sin\theta_{t-1,t}
    ].
\end{equation}
The absolute ending position of cell-flow edge $(v_{t-1},v_t)$ is the summation of the position vectors through the cell-path. As there might be multiple cell-flow edges ending at cell $v_t$, the decoded absolute position of cell $v_t$ is averaged over edges:
\begin{equation}
    \vp_{t;(t-1,t)}=\vp_0+\sum_{i=1,\cdots,t}\vec\vp_{i-1,i}\qquad
    \vp_t=\underset{(v_i,v_t)\in\sF}{\rm mean}\vp_{t;(i,t)}.
    \label{eq:dec-2}
\end{equation}
Overall, by stacking $\vp_j,j\in\sV_M$, we obtain the decoding results:
\begin{equation}
    \mP_M=f_{\rm DECODE}(\bm{\rho},\Delta\vtheta;\gG).
    \label{eq:decode}
\end{equation}

The time consumption of producing the position vectors for all edges in $\sF$ is $\omega|\sF|$, where $\omega$ is the average length of cell-paths. Since $\omega$ is typically small (see \cref{sec:opt-coding}), it can be treated as a constant. We also have $|\sF|=O(|\sP|)$ (\cref{sec:opt-flow}). As a result, the time required for relative position decoding is proportional to the scale of the netlist graph, which is optimal.

\subsection{\gnn{}}
\label{sec:gnn}
\subsubsection{Inference}
We illustrate the inference pipeline of \gnn{} in \cref{fig:method}. \gnn{} takes a netlist graph and its cell-flow as inputs, then embeds raw features $\mX_{\sV}$, $\mX_{\sU}$, and $\mX_{\sP}$ into hidden representations $\mH_{\sV}^{(0)}$, $\mH_{\sU}^{(0)}$, $\mH_{\sP}$, via Multi-Layer Perceptrons (MLPs). Then, it generates deeper representations of cells $\sV$, and nets $\sU$ with $L$ layers of message-passing. Finally, readout layers convert the output cell and net representations into relative cell positions.

In each layer, the topological information is collected through cell-flow and netlist graph message-passing. The messages are then used to fuse and update the representations of cells $\sV$ and nets $\sU$. For netlist graph message-passing, we interact the messages of cells $\sV$ and nets $\sU$ through graph edges which preserve the topological connection between cells and nets:
\begin{align}
    \mM_{\sU\rightarrow\sV}^{(l)}=
f_{\sU\rightarrow\sV}(\sU,\sP,\mH_\sU^{(l)},\mH_\sP), \\
\mM_{\sV\rightarrow\sU}^{(l)}=
f_{\sV\rightarrow\sU}(\sV,\sP,\mH_\sV^{(l)},\mH_\sP),
\end{align}
where $l$ is the number of current layer, $\mM_{\sV}^{(l)}$ and $\mM_{\sU}^{(l)}$ denote the messages of cells $\sV$ and nets $\sU$ computed on netlist graphs. $f_{\sV\rightarrow\sU}$ is the message function that collects topological messages from nets $\sU$ and sends them to cells $\sV$ via netlist graph edges. $f_{\sU\rightarrow\sV}$ is similar. $\mH_{\sV}^{(l)}$ and $\mH_{\sU}^{(l)}$ denote the hidden representations of the cells $\sV$ and nets $\sU$ of layer $l$. Inspired by SchNet \cite{schnet} we design the message function $f_{\sU\rightarrow\sV}$ to fuse representations of adjacent cells and edges. Similarly, influenced by CircuitGNN \cite{routeplacer}, we design $f_{\sV\rightarrow\sU}$ to utilize edge representations for computing weights during message passing, enabling cells to perceive geometric information. These functions can be expressed as follows:
\begin{align}
    f_{\sU\rightarrow\sV}(\{(\vh^\sU_u,\vh^\sP_{(v,u)})|(v,u)\in\sP\})=\nonumber\\
	\sum_{(v,u)\in\sP}
	(\mW_{\sP\rightarrow\sV}\vh^\sP_{(v,u)})\odot
	\mW_{\sU\rightarrow\sV}\vh^\sU_u \\
 f_{\sV\rightarrow\sU}(\{(\vh^\sV_v,\vh^\sP_{(v,u)})|(v,u)\in\sP\})=\nonumber\\
	\sum_{(v,u)\in\sP}
	(\va^\top\vh^\sP_{(v,u)})\cdot
	(\mW_{\sV\rightarrow\sU}\vh^\sV_v),
\end{align}
where $\mW_{\sP\rightarrow\sV}\vh^\sP_{(v,u)}$, $\mW_{\sU\rightarrow\sV}\vh^\sU_u$, and $\mW_{\sV\rightarrow\sU}\vh^\sV_v$ are learnable weight matrices; $\va$ is a learnable vector; and $\odot$ refers to inner-product.

For Cell-Flow message-passing, we directly collect messages of cells $\sV$ to obtain a fused message:
\begin{align}
     \mM_{\sV\rightarrow\sV}^{(l)}=
	f_{\sV\rightarrow\sV}(\sV,\sF,\mH_\sV^{(l)}).
\end{align}
Here, $\mM_{\sV\rightarrow\sV}^{(l)}$ denotes the messages of cells $\sV$ computed on cell-flow and $\mH_\sV^{(l)}$ denotes the hidden representations of cells $\sV$. $f_{\sV\rightarrow\sV}$ is the message function that transfers relative positional relationship messages from cells $\sV$. We design $f_{\sV\rightarrow\sV}$ to speed up the message-passing as below:
\begin{align}
f_{\sV\rightarrow\sV}(\{\vh^\sV_v|(v,v^*)\in\sF\})=
	\sum_{(v,v^*)\in\sF}
	\mW_{\sV\rightarrow\sV}\vh^\sV_{v^*},
\end{align}
where $\mW_{\sV\rightarrow\sV}\vh^\sV_{v^*}$ is a learnable weight matrix.

After computing the netlist graph and cell-flow messages for cells $\sV$, we update the representations of cells $\sV$ with the fused messages $\mM_{\sV}^{(l)}$:
\begin{align}
    \mM_{\sV}^{(l)} = MaxPooling(\mM_{\sV\rightarrow\sV}^{(l)},\mM_{\sU\rightarrow\sV}^{(l)}).
\end{align}
At last, we combine the messages and hidden representations from layer $l$ to generate the updated hidden representations for layer $l+1$:
\begin{align}
    \mH_\sV^{(l+1)}=f_{\rm update}(\mH_\sV^{(l)}, \mM_{\sV}^{(l)}) \\
	\mH_\sU^{(l+1)}=f_{\rm update}(\mH_\sU^{(l)},\mM_{\sV\rightarrow\sU}^{(l)})
\end{align}
Here, the update function $f_{\rm update}(\mH,\mM)=\mH+\mM$.

After $L$ iterations of message passing, we obtain the final representation of cells $\mH_\sV=\mH_\sV^{(L)}$ and nets $\mH_\sU=\mH_\sU^{(L)}$, and readout the relative cell positions, i.e., distance $\hat \rho$ and deflection $\Delta \hat \theta$ of cell flow edges $(v_i, v_j) \in \sF$:
\begin{align}
	\vh^\rho_{i,j} &= \vh^\sV_i\oplus\vh^\sU_t\oplus\vh^\sV_j, 
	\\
	\vh^\theta_{i,j}&=\vh^\sV_k\oplus\vh^\sU_s\oplus
	\vh^\sV_i\oplus\vh^\sU_t\oplus\vh^\sV_j,
\end{align}
where $\oplus$ means concatenation, and $u_s=f_{\rm NET}((v_k,v_i))$, $u_t=f_{\rm NET}(($ $v_i,v_j))$, $(v_k,v_i)=f_{\rm INCOME}(v_i)$. Then we define:
\begin{align}
	\hat \rho_{i,j} = f_\rho(\vh^\rho_{i,j})=\exp(\beta+\alpha\cdot
	{\rm Tanh}(\va_\rho^\top\cdot\vh^\rho_{i,j}+b_\rho)) \\
 \Delta \hat \theta_{i,j} = f_\theta(\vh^\theta_{i,j})=\pi\cdot
	{\rm Tanh}(\va_\theta^\top\cdot\vh^\theta_{i,j}+b_\theta),
\end{align}
where $\va_\rho$, $\va_\theta$, $b_\rho$, and $b_\theta$ are learnable weight vectors and scalars. $\alpha$ and $\beta$ are hyperparameters used to control the range of $\rho$. In this paper, we set $\alpha=15$ and $\beta=-2$ to flexibly adapt to $\rho\in(4e-8, 4e5)$.

\subsubsection{Training}

\gnn{} is trained to imitate preplaced circuits generated by DREAMPlace \cite{dreamplace}, i.e., netlist graphs $\gG$ with absolute positions for movable cells $\hat\mP_M$. We encode the absolute positions into ground-truth relative ones as training labels.
The training loss is determined by the difference between \gnn{} outputs and ground truth relative positions using ${\rm Smooth\text{-}L1}$ loss. We defer the full details to \cref{sec:app-gnn-training}.

\subsection{Circuit-adaptive Fine-tuning}\label{sec:fine-tuning}

While inductively generating placement is efficient, it may neglect the unique characteristics of an unseen circuit, such as its terminal positions and core placement area. Therefore, after generating the inductive placements, we iteratively fine-tune them by minimizing wirelength \cite{tsv} and density \cite{eplace}.

The wirelength objective is calculated given cell positions \cite{tsv}:
\begin{align}
    \gL_{\rm W}=\gL_{\rm W}(x)+\gL_{\rm W}(y)&\\
	\gL_{\rm W}(x)=\sum_{u\in\sU}(
	\frac{\sum_{(v,u)\in\sP}x_v e^{\gamma x_v}}{\sum_{(v,u)\in\sP}e^{\gamma x_v}}-&\nonumber \\
	\frac{\sum_{(v,u)\in\sP}x_v e^{-\gamma x_v}}{\sum_{(v,u)\in\sP}e^{-\gamma x_v}}&
	),
 \label{eq:ft-wl}
\end{align}
where $x_v$ denotes the x-coordinate of cell $v$ and $\gamma$ is a hyperparameter. $\gL_{\rm W}(y)$ is computed in a similar manner.

The unique solution of the electrostatic system is derived from \cite{eplace}:
\begin{align}
	\left\{
	\begin{array}{l}
		\nabla\cdot\nabla\psi(x,y)=-\rho(x,y)\\
		\hat n\cdot\nabla\psi(x,y)=0,(x,y)\in\partial R\\
		\iint_R\rho(x,y)=\iint_R\psi(x,y)=0
	\end{array}
	\right. \\
 \gL_{\rm D}=\sum_{i\in\sV}N_i(v)=\sum_{i\in\sV}q_i\psi(v),\label{eq:ft-density}
\end{align}
where $\rho(x,y)$ denotes the cell density in position $(x,y)$ and $\psi(x,y)$ is the potential in position $(x,y)$ \cite{eplace}.

Overall, the fine-tuning loss is given by \cref{eq:fine-tune-obj}, where $\lambda_{\rm D}$ weighs two loss terms. The fine-tuning parameters are updated after backward propagation and cell position updates, according to the rules given below.
\begin{align}
    \lambda_D&=\lambda_D*\mu\\
    \mu &=\left\{ {\begin{smallmatrix}
		1.05*\max\mathrm{(}0.999^{epochs},0.98)&		\Delta HPWL<0\\
		1.05*1.05^{\frac{-\Delta HPWL}{350000}}&		\Delta HPWL\geqslant 0\\
	\end{smallmatrix} } \right. \\
	\Delta HPWL&=HPWL_{new}-HPWL_{old} \\
 \gL_{\rm fine-tune}&=\gL_{\rm W}+\lambda_{\rm D}\gL_{\rm D}. \label{eq:fine-tune-obj}
\end{align}
Here, HPWL is the half-perimeter wire length \cite{dreamplace}. $HPWL_{new}$ and $HPWL_{old}$ represent the HPWL before and after this optimization step, respectively.


\section{Experiments}

This section comprehensively evaluates \ours{} and answers two key research questions (RQs): 

\begin{enumerate}
    \item Can \ours{} transfer placement knowledge and enhance diverse unseen circuit designs? 
    \item Can \ours{} improve multiple design objectives simultaneously \cite{cross-obj, power-cross-obj}, even when primarily trained with specific focuses like congestion?
\end{enumerate}


\begin{table*}[htbp]
	\caption{\centering{Comparison results on \texttt{ISPD2015}.}}
	\label{tab:RQ1-2}
\centering
\begin{tabular}{cccccccccc}
 
		\toprule
		\multirow{2}{*}{Benchmark} &
		\multicolumn{1}{c}{\multirow{2}{*}{Netlist}} &
		\multicolumn{1}{c}{\multirow{2}{*}{\#cell}} &
		\multicolumn{1}{c}{\multirow{2}{*}{\#nets}} &
		\multicolumn{3}{c}{DREAMPlace} &
		\multicolumn{3}{c}{\ours{}} \\ \cmidrule{5-10} 
		&
		\multicolumn{1}{c}{} &
		\multicolumn{1}{c}{} &
		\multicolumn{1}{c}{} &
		RT &
		OVFL$\downarrow$ &
		RWL($\times10^6um$)$\downarrow$ &
		RT &
		OVFL &
		RWL \\ \midrule
		\multirow{17}{*}{ISPD2015} &
		mgc\_des\_perf\_1 &
		113K &
		113K &
		\textbf{34.21} &
		41 &
		\textbf{1.70} &
		37.25 &
		\textbf{24} &
		1.81 \\
		&
		mgc\_des\_perf\_a &
		109K &
		110K &
		143.58 &
		13,123 &
		\textbf{3.06} &
		\textbf{71.25} &
		\textbf{11,841} &
		3.22 \\
		&
		mgc\_des\_perf\_b &
		113K &
		113K &
		107.58 &
		13 &
		\textbf{2.38} &
		\textbf{90.01} &
		\textbf{7} &
		2.59 \\
		&
		mgc\_edit\_dist\_a &
		130K &
		131K &
		65.93 &
		16,251 &
		6.18 &
		\textbf{61.53} &
		\textbf{13,738} &
		\textbf{6.15} \\
		&
		mgc\_fft\_1 &
		35K &
		33K &
		35.14 &
		19 &
		\textbf{0.65} &
		\textbf{11.05} &
		\textbf{5} &
		0.70 \\
		&
		mgc\_fft\_2 &
		35K &
		33K &
		41.37 &
		5 &
		\textbf{0.65} &
		\textbf{26.18} &
		\textbf{2} &
		0.72 \\
		&
		mgc\_fft\_a &
		34K &
		32K &
		35.24 &
		3,244 &
		\textbf{0.99} &
		\textbf{11.64} &
		\textbf{2,604} &
		1.01 \\
		&
		mgc\_matrix\_mult\_1 &
		160K &
		159K &
		37.51 &
		9 &
		\textbf{3.01} &
		\textbf{21.96} &
		\textbf{5} &
		3.03 \\
		&
		mgc\_matrix\_mult\_2 &
		160K &
		159K &
		39.60 &
		12 &
		3.00 &
		\textbf{20.54} &
		\textbf{0} &
		\textbf{2.98} \\
		&
		mgc\_matrix\_mult\_a &
		154K &
		154K &
		32.69 &
		6,802 &
		4.02 &
		\textbf{18.54} &
		\textbf{4,477} &
		\textbf{3.91} \\
		&
		mgc\_pci\_bridge32\_a &
		30K &
		30K &
		157.47 &
		3,376 &
		\textbf{0.77} &
		\textbf{156.17} &
		\textbf{2,776} &
		0.77 \\
		&
		mgc\_pci\_bridge32\_b &
		29K &
		29K &
		216.42 &
		2,921 &
		\textbf{1.01} &
		\textbf{147.57} &
		\textbf{2,419} &
		1.01 \\
		&
		mgc\_superblue11\_a &
		954K &
		936K &
		97.39 &
		899 &
		\textbf{44.04} &
		\textbf{85.69} &
		\textbf{561} &
		44.46 \\
		&
		mgc\_superblue12 &
		1293K &
		1293K &
		\textbf{70.33} &
		99,641 &
		\textbf{40.86} &
		96.47 &
		\textbf{67,690} &
		40.98 \\
		&
		mgc\_superblue14 &
		634K &
		620K &
		35.33 &
		1,690 &
		\textbf{29.54} &
		\textbf{33.46} &
		\textbf{1,283} &
		30.05 \\
		&
		mgc\_superblue16\_a &
		698K &
		697K &
		\textbf{82.64} &
		27,862 &
		34.01 &
		115.74 &
		\textbf{26,894} &
		\textbf{33.59} \\
		&
		mgc\_superblue19 &
		522K &
		512K &
		\textbf{27.23} &
		432 &
		\textbf{20.47} &
		30.81 &
		\textbf{110} &
		20.76 \\
		\multicolumn{1}{l}{} &
		Average ratio &
		\multicolumn{1}{c}{} &
		\multicolumn{1}{c}{} &
		1.50 &
		1.31 &
		0.98 &
		1.00 &
		1.00 &
		1.00 \\ \bottomrule
\end{tabular}
\end{table*}

\begin{table*}[htbp]
	\caption{\centering{Comparison results on \texttt{ISPD2019}.}}
	\label{tab:RQ1-3}
\centering
	\begin{tabular}{cccccccccc}
 
		\toprule
		\multirow{2}{*}{Benchmark} &
		\multicolumn{1}{c}{\multirow{2}{*}{Netlist}} &
		\multicolumn{1}{c}{\multirow{2}{*}{\#cell}} &
		\multicolumn{1}{c}{\multirow{2}{*}{\#nets}} &
		\multicolumn{3}{c}{DREAMPlace} &
		\multicolumn{3}{c}{\ours{}} \\ \cmidrule{5-10} 
		&
		\multicolumn{1}{c}{} &
		\multicolumn{1}{c}{} &
		\multicolumn{1}{c}{} & 
		RT &
		OVFL$\downarrow$ &
		RWL($\times10^6um$)$\downarrow$ &
		RT &
		OVFL &
		RWL \\ \midrule
		\multicolumn{1}{l}{\multirow{9}{*}{ISPD2019}} &
		ispd19\_test1 &
		9K &
		3K &
		9.89 &
		0 &
		0.09 &
		\textbf{9.56} &
		0 &
		\textbf{0.09} \\
		\multicolumn{1}{l}{} &
		ispd19\_test2 &
		73K &
		72K &
		\textbf{40.75} &
		0 &
		\textbf{2.72} &
		42.61 &
		0 &
		2.72 \\
		\multicolumn{1}{l}{} &
		ispd19\_test3 &
		8K &
		9K &
		\textbf{15.69} &
		0 &
		0.12 &
		18.25 &
		0 &
		\textbf{0.12} \\
		\multicolumn{1}{l}{} &
		ispd19\_test4 &
		151K &
		152K &
		32.88 &
		9299 &
		\textbf{4.68} &
		\textbf{23.26} &
		\textbf{9159} &
		4.69 \\
		\multicolumn{1}{l}{} &
		ispd19\_test6 &
		181K &
		180K &
		53.27 &
		0 &
		\textbf{6.92} &
		\textbf{44.41} &
		0 &
		6.98 \\
		\multicolumn{1}{l}{} &
		ispd19\_test7 &
		362K &
		359K &
		55.30 &
		0 &
		\textbf{13.66} &
		\textbf{46.95} &
		0 &
		14.39 \\
		\multicolumn{1}{l}{} &
		ispd19\_test8 &
		543K &
		538K &
		58.26 &
		0 &
		\textbf{20.59} &
		\textbf{49.32} &
		0 &
		20.94 \\
		\multicolumn{1}{l}{} &
		ispd19\_test9 &
		903K &
		895K &
		\textbf{62.85} &
		0 &
		\textbf{32.75} &
		70.65 &
		0 &
		33.47 \\
		\multicolumn{1}{l}{} &
		ispd19\_test10 &
		903K &
		895K &
		68.83 &
		366 &
		\textbf{33.21} &
		\textbf{54.91} &
		\textbf{143} &
		35.46 \\
		\multicolumn{1}{l}{} &
		Average ratio &
		\multicolumn{1}{c}{} &
		\multicolumn{1}{c}{} &
		1.06 &
		2.43 &
		0.98 &
		1.00 &
		1.00 &
		1.00 \\ \bottomrule
\end{tabular}
\end{table*}

\subsection{Knowledge Transfer across Diverse Designs}
\label{sec:transfer-circuits}

For the first RQ, our evaluations involve three benchmarks containing 37 circuits with diverse functionalities. We pick 5 large-scale circuits as training datasets to ensure sufficient transferable placement knowledge. To compare our model with other global placers, we use overflow and routed wirelength, measured after global routing, as our evaluation metrics. Although traditional placer usually takes half perimeter wirelength (HPWL) as an evaluation metric, it does not consider detours in the path, and wire congestion \cite{routing} may still cause routing failure even for solutions with better HPWL. Therefore, we evaluate placements after global routing to provide an accurate evaluation of routability quality. We conduct the following steps to compute overflow. We first divide the entire layout into grids, each with a predefined wire capacity denoted as $RC$. This limit represents the maximum number of wires allowed in each grid cell. Overflow $OF(i,j)$ occurs when the number of wires exceeds this limit in the grid cell $(i,j)$. Total overflow can be defined as follows:
\begin{align}
    TOF=\sum_{i,j}OF(i,j).
\end{align}

\begin{table*}[htbp]
	\caption{\centering{Comparison results on \texttt{DAC2012}}}
	\label{tab:RQ1-1}
\centering
 \resizebox{\linewidth}{!}{
	\begin{tabular}{ccccccccccccc}
		\toprule
		\multirow{2}{*}{Benchmark} &
		\multicolumn{1}{c}{\multirow{2}{*}{Netlist}} &
		\multicolumn{1}{c}{\multirow{2}{*}{\#cell}} &
		\multicolumn{1}{c}{\multirow{2}{*}{\#nets}} &
		\multicolumn{3}{c}{NTUPlace} &
		\multicolumn{3}{c}{DREAMPlace} &
		\multicolumn{3}{c}{\ours{}} \\ \cmidrule{5-13} 
		&
		\multicolumn{1}{c}{} &
		\multicolumn{1}{c}{} &
		\multicolumn{1}{c}{} &
		RT &
		OVFL$\downarrow$ &
		RWL($\times10^6um$)$\downarrow$ &
		RT &
		OVFL &
		RWL &
		RT &
		OVFL &
		RWL \\ \midrule
		\multirow{10}{*}{DAC2012} &
  superblue2 &
		1014K &
		991K &
		6028 &
		115624 &
		23.65 &
		337.17 &
		54220 &
		24.37 &
		\textbf{158.80} &
		\textbf{33808} &
		\textbf{23.66} \\
		&
		superblue3 &
		920K &
		898K &
		5443 &
		212946 &
		16.78 &
		\textbf{174.25} &
		7574 &
		\textbf{15.43} &
		202.70 &
		\textbf{6432} &
		15.56 \\
		&
		superblue6 &
		1014K &
		1007K &
		5374 &
		143118 &
		16.23 &
		248.75 &
		4092 &
		15.78 &
		\textbf{219.97} &
		\textbf{2090} &
		\textbf{15.32} \\
		&
		superblue7 &
		1365K &
		1340K &
		8406 &
		4780 &
		20.94 &
		183.39 &
		5182 &
		20.15 &
		\textbf{148.69} &
		\textbf{4794} &
		\textbf{20.15} \\
		&
		superblue9 &
		847K &
		834K &
		6039 &
		41942 &
		12.01 &
		128.59 &
		3700 &
		12.40 &
		\textbf{115.70} &
		\textbf{2644} &
		\textbf{12.12} \\
		&
		superblue11 &
		955K &
		936K &
		5621 &
		18342 &
		15.45 &
		\textbf{167.64} &
		16606 &
		17.44 &
		197.41 &
		\textbf{2856} &
		\textbf{16.14} \\
		&
		superblue12 &
		1293K &
		1293K &
		27751 &
		580406 &
		17.67 &
		80.30 &
		21804508 &
		36.59 &
		\textbf{59.36} &
		\textbf{2777526} &
		\textbf{17.73} \\
		&
		superblue14 &
		635K &
		620K &
		4368 &
		10510 &
		10.71 &
		143.81 &
		19284 &
		\textbf{10.41} &
		\textbf{99.38} &
		\textbf{13552} &
		10.56 \\
		&
		superblue16 &
		699K &
		697K &
		5089 &
		16618 &
		11.60 &
		137.78 &
		10148 &
		11.03 &
		\textbf{96.38} &
		\textbf{9422} &
		\textbf{10.87} \\
		&
		superblue19 &
		523K &
		512K &
		3749 &
		33546 &
		7.48 &
		\textbf{73.89} &
		7422 &
		7.29 &
		83.17 &
		\textbf{3478} &
		7.40 \\
		\multicolumn{1}{l}{} &
		Average ratio &
		\multicolumn{1}{c}{} &
		\multicolumn{1}{c}{} &
		83.57 &
		14.06 &
		1.02 &
		1.24 &
		2.55 &
		1.12 &
		1.00 &
		1.00 &
		1.00 \\ \bottomrule
\end{tabular}
}
\end{table*}

\begin{table*}[htp]
	\caption{\centering{Comparison results on \texttt{ICCAD2015}}.}
	\label{tab:RQ2}
 \centering
 \resizebox{\linewidth}{!}{
		\begin{tabular}{lcccccccccc}
			\toprule
			\multicolumn{1}{c}{\multirow{2}{*}{netlist\_name}} &
			\multirow{2}{*}{\#cell} &
			\multirow{2}{*}{\#nets} &
			\multicolumn{4}{c}{DREAMPlace} &
			\multicolumn{4}{c}{\ours{}} \\ \cmidrule{4-11}
			\multicolumn{1}{c}{} &
			&
			&
			WNS$\downarrow$ &
			TNS$\downarrow$ &
			NVP$\downarrow$ &
			rWL$\downarrow$ &
			WNS &
			TNS &
			NVP &
			rWL \\ \midrule
			superblue1 &
			1216K &
			1215K &
			\textbf{-16.98} &
			-27864.90 &
			\textbf{7922} &
			\textbf{99590980} &
			-20.10 &
			\textbf{-25575.90} &
			\textbf{7971} &
			\textbf{100961600} \\
			superblue3 &
			1219K &
			1224K &
			-17.48 &
			-12899.00 &
			4412 &
			98362130 &
			\textbf{-14.86} &
			\textbf{-11608.30} &
			\textbf{3700} &
			\textbf{97555240} \\
			superblue4 &
			802K &
			802K &
			-20.81 &
			\textbf{-24490.30} &
			\textbf{5967} &
			\textbf{67779710} &
			\textbf{-16.41} &
			-30316.90 &
			8493 &
			74526630 \\
			superblue5 &
			1091K &
			1100K &
			-25.31 &
			-20993.40 &
			8094 &
			107397500 &
			\textbf{-25.17} &
			\textbf{-12956.90} &
			\textbf{4708} &
			\textbf{99634140} \\
			superblue7 &
			1938K &
			1933K &
			\textbf{-16.40} &
			\textbf{-13619.30} &
			\textbf{7683} &
			\textbf{125085700} &
			\textbf{-14.68} &
			\textbf{-13105.80} &
			8415 &
			126574700 \\
			superblue10 &
			1888K &
			1898K &
			\textbf{-29.40} &
			\textbf{-103000.00} &
			\textbf{12193} &
			217113500 &
			-33.38 &
			-105000.00 &
			13122 &
			\textbf{202172100} \\
			superblue16 &
			986K &
			999K &
			\textbf{-14.50} &
			-25465.30 &
			11591 &
			\textbf{91522550} &
			\textbf{-13.65} &
			\textbf{-21145.00} &
			\textbf{10747} &
			93100710 \\
			superblue18 &
			772K &
			771K &
			-17.50 &
			\textbf{-11082.30} &
			\textbf{3336} &
			49204750 &
			\textbf{-13.93} &
			-13300.60 &
			5663 &
			\textbf{48455550} \\
			Average ratio&
			&
			&
			1.08 &
			1.09 &
			1.01 &
			1.01 &
			1.00 &
			1.00 &
			1.00 &
			1.00 \\ \bottomrule
	\end{tabular}
 }
	\label{tab:iccad2015}
\end{table*}

We evaluate \ours{} against two classical placers: (1) NTUPlace \cite{NTUPlace4h}, an analytical placer for mixed-size circuit designs, and (2) DREAMPlace \cite{dreamplace}, recognized for its efficient implementation of GPU acceleration in analytical placement. The global router NCTUgr \cite{NCTUgr} is used for the \texttt{DAC2012} dataset \cite{DAC2012}, while FastRoute 4.0 \cite{FastRoute4.0} for the \texttt{ISPD2015} and \texttt{ISPD2019} datasets \cite{ispd2015, ISPD2019}. Detailed placement is conducted using Abacus \cite{abacus} and ABCDPlace \cite{abcdplace} for final placement results. For NTUPlace, we set the number of maximum threads to 8. All the experiments are conducted using a single Nvidia RTX 3090 GPU and an Intel Platinum 8255C CPU.

The results on three datasets are shown in \cref{tab:RQ1-2}, \cref{tab:RQ1-3}, and \cref{tab:RQ1-1}.
The evaluation encompasses the aggregate routed wirelength, the total overflow, and the total runtime including placement and fine-tuning steps which are short for RWL, OVFL, and RT. 
As shown by the results, compared to the state-of-the-art placer DREAMPlace, \ours{} can place a new circuit about 1.2x faster, with a 30\% reduction in Total Overflow and 2\% reduction on Wirelength (averaged over circuits in \texttt{ISPD2015} and \texttt{ISPD2019}). A lower total overflow not only implies a reduced likelihood of wire congestion, but also enhances the potential for successful routability and design convergence. Notably, despite our training dataset comprising only five circuits from \texttt{DAC2012}, \ours{} exhibits adaptability across a wide array of circuits from three distinct benchmarks. This adaptability is important for meeting the ever-evolving demands of chip design. It demonstrates the potential of \ours{} to improve EDA workflows in practical, high-stakes environments.

\subsection{Cross-Objective Optimization}

\begin{figure*}
	\includegraphics{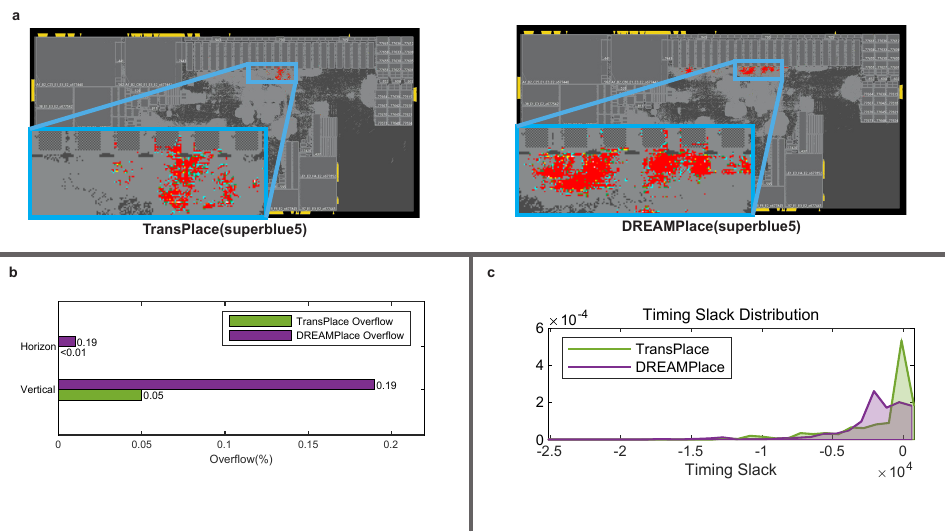}
	\caption{Visualizations of comparisons on \texttt{superblue5}. (a) Congestion visualization of the placement generated by DREAMPlace and \ours{}. The grey parts denote the cells, and the red dots indicate that there is congestion. Density of red dots indicates the level of congestion. (b) Overall comparison of vertical and horizontal overflow (lower overflow implies better performance). (c) Distribution of timing path slack for placements. A higher proportion of smaller slack indicates better performance.
 }
 \Description{}
	\label{fig:RQ2}
\end{figure*}

\ours{} is trained on placements generated by DREAMPlace with routability-driven methods. To answer the second RQ, this section evaluates its performance on the \texttt{ICCAD2015} dataset \cite{iccad2015}, which focuses on incremental timing-driven placement \cite{dreamplace4, fpga-timing}. This placement approach optimizes the positions of circuit elements to minimize interconnect delay in timing-critical paths. The metrics used for the evaluation are total negative slack (TNS), worst negative slack (WNS), and number of violation paths (NVP).

The calculation of TNS, WNS, and NVP is based on the timing graph, a directed acyclic graph defined by the circuits. In this graph, each node represents a pin in the circuit, and each edge indicates a directed pin-to-pin connection. We travel all nodes connected by edges in the graph, beginning at the source cell node. For each node, we calculate its actual arrival time; the difference between it and the required arrival time is termed "slack" \cite{slack}. The maximum slack across all nodes is known as the Worst Negative Slack (WNS), while the Total Negative Slack (TNS) is the sum of the slacks at all timing endpoints. A path with negative slack at any of its nodes is referred to as a violation path, and NVP denotes the total number of such paths.

We compare \ours{} with DREAMPlace 4.0 \cite{dreamplace4}, DREAMPlace with timing-driven methods. In our Circuit-adaptive Fine-tuning approach, similar to DREAMPlace 4.0, we employ OpenTimer for evaluating TNS and WNS and adopt a momentum-based method for updating net weights used in the wirelength calculation. For the assessment of our timing metrics, which include TNS, WNS, and NVP, we utilize Cadence Innovus. Detailed placement is conducted using Abacus \cite{abacus} and ABCDPlace \cite{abcdplace} for final placement results.

\cref{tab:RQ2} presents the complete comparison results on \texttt{ICCAD2015} benchmark, where the best results are highlighted in bold and black. As shown by the results, compared to DREAMPlace, \ours{} can place a new circuit with a 9\% reduction on TNS, 1\% reduction on NVP, 8\% reduction on WNS, and 1\% reduction on rWL (averaged over circuits in the dataset). Although \ours{} uses routability-driven placement as the training labels, it shows improvement in timing metrics for unseen circuits. This capability evidences the model's proficiency in extending learned knowledge beyond the initial training focus. 
The ability to generalize from one objective to another ensures an adaptable and efficient design process, especially when dealing with complex circuits. \ours{} allows designers to improve multiple aspects simultaneously, showing promise to streamline the overall design process.

\cref{fig:RQ2} visualizes the comparisons between DREAMPlace and \ours{} on \texttt{superblue5}. \cref{fig:RQ2}(a) reveals that our solution achieves more efficient congestion management, evidenced by a discernible reduction in red dots and fewer congestion peaks. \cref{fig:RQ2}(b) indicates our solution achieves much less vertical and horizontal overflow. Furthermore, \cref{fig:RQ2}(c) displays the distribution in timing path slack of \ours{} and DREAMPlace, which shows overall optimization for timing slack. 

These findings emphasize the technical advances \ours{} provides and its applicability in real-world chip design scenarios. By effectively reducing congestion and optimizing timing path slack, \ours{} addresses the multifaceted challenges of designing contemporary complex circuits. This capability to simultaneously optimize multiple design objectives highlights \ours{}'s potential to streamline the chip design process.

\section{Related Work}
\paragraph{Global Placement.}
Global placement methods have been extensively developed since the 1960s. The proposed methods are broadly classified into four categories: meta-heuristics \cite{mdptree, hidap, evolv, ye2024reevo}, partition-based methods \cite{min-cut}, RL-based methods \cite{googleplace, joint, morl, chipformer, macro-mask, berto2024rl4co}, and analytical methods \cite{eplace, dreamplace, replace, ripple, ntuplace3}.
Meta-heuristics treat placement as a step-wise optimization problem and solve it with heuristic algorithms such as Simulated Annealing and Genetic Algorithm. Partition-based methods iteratively divide the chip's netlist and layout into smaller sub-netlists and sub-layouts, based on the cost function of the cut edges. Optimization methods are used to find solutions when the netlist and layout are sufficiently small. RL-based methods regard the placement problem as a Markov Decision Process (MDP) and train a policy to sequentially place the cells. All of them suffer from low convergence rates, which restricts their usage only to the circuits with a small number of large-sized cells. 
Compared to them, analytical methods can optimize modern circuits with millions of cells. These methods optimize the positions of mixed-size cells by formulating the problem as an analytical function of cell coordinates. Analytical methods have gained popularity recently due to their compatibility with deep learning toolkits, which enables GPU-accelerated placements \cite{elfplace}. However, they approach each placement instance independently from scratch, unable to utilize previous placement experiences. 
Prior to our work, the complexity and scale of global placement hindered the application of deep learning techniques for learning-based enhancement.

\paragraph{Graph Neural Networks (GNNs) for Placement.}
GNNs demonstrate state-of-the-art performance in predictive and generative tasks involving graph data \cite{wu2020gnn_survey}. Since integrated circuits (ICs) can naturally be represented as graphs, there has been a growing adoption of GNNs in IC design, particularly in the critical placement phase \cite{alrahis2023gnn_ic}.
GNNs are applied to various tasks in the EDA workflow, including behavioral and logic design \cite{wu2021ironman, lopera2022applying}, logic synthesis \cite{kirby2019congestionnet, zhou2022heterogeneous}, partitioning and floorplanning \cite{googleplace, chipformer, jiang2021delving, cheng2021joint, cheng2022policy, liu2022floorplanning, liu2022graphplanner, lai2022maskplace, shi2024macro}, routing congestion prediction \cite{lhnn, routeplacer}, timing analysis \cite{lu2021doomed, lu2021rl}, etc. Among them, most relevant to our work is the application of GNNs for the floorplanning stage. However, they still resort to optimization-based methods for global placement due to their limited scalability. Prior to this work, the application of GNNs to inductive global placement remains limited due to significant technical challenges (\cref{sec:introduction}). 

\section{Conclusion}
\label{sec:conclusion}


This work introduces \ours{}, the first learning-based approach for global placement in integrated circuit design. We develop a series of techniques to address the technical challenges of learning large-scale cell placement. We demonstrate that \ours{} can effectively transfer versatile placement knowledge through evaluations on four benchmarks that feature diverse circuits and design objectives.

\ours{} shows promise in improving integrated circuit design by learning parameterized knowledge, which results in higher-quality circuits and accelerates design cycles. \ours{} can serve as an initial setup for placers, streamlining the optimization process for customized objectives and providing warm starts for optimization. Additionally, for designs requiring multiple logic optimizations, \ours{} can efficiently provide placement solutions, enabling early evaluation and issue resolution, thereby speeding up design convergence.

\section{ACKNOWLEDGMENTS}
\label{sec:ackowledgments}
This work was supported by the National Natural Science Foundation of China (Grant No. 62276006).

\bibliographystyle{ACM-Reference-Format}
\balance
\bibliography{document}

\appendix
\section{Sub-netlist Graph Construction}
\label{sec:app-sub-netlist}

The construction of sub-netlist graph follows \cref{alg:hier}.

\begin{breakablealgorithm}
  \caption{Extracting sub-netlist graphs and coarsening original netlist graph.}
  \label{alg:hier}
\begin{algorithmic}[1]
  \STATE {\bfseries Input:} netlist graph $\gG=\{\sV,\sU,\sP,\mX_\sV,\mX_\sU,\mX_\sP,\mP_T,\mP_P\}$, partition sets $\{\sV_i\}$
  
  \STATE Initialize $\sU_i=\emptyset,\sP_i=\emptyset\quad\forall i=1,\cdots,n$
  \STATE Initialize $f_{\rm BELONG}$ with $\{\sV_i\}$
  \FORALL{$(v,u)\in\sP\ \text{with}\ f_{\rm BELONG}(v)\ne0$}
  \STATE Add $u$ into $\sU_{f_{\rm BELONG}(v)}$
  \STATE Add $(v,u)$ into $\sP_{f_{\rm BELONG}(v)}$
  \ENDFOR

  \STATE Initialize pseudo cell set $\hat\sV_M=\emptyset$
  \FORALL{$i\in\{1,\cdots,n\}$}
  \STATE Initialize feature matrices:
  \begin{align}
      \mX_{\sV,i}=\mX_\sV[\sV_i,:],
    \mX_{\sU,i}=\mX_\sU[\sU_i,:],
    \mX_{\sP,i}=\mX_\sP[\sP_i,:]
  \end{align}
    
  \STATE Find the biggest cell $v^*\in\sV_i$
  \STATE Initialize terminal and movable:
  \begin{equation}
      \sV_{T,i}=\{v^*\},\sV_{P,i}=\emptyset,\sV_{M,i}=\sV_i/v^*
  \end{equation}
  \STATE Initialize cell positions $\mP_{T,i}=[[0,0]],\mP_{P,i}=[]$
  \STATE Construct $i$-th sub-netlist graph:
  \begin{align}
      \gG_i=\{\sV_{T,i}\cup\sV_{M,i},\sU_i,\sP_i,
    \mX_{\sV,i},\mX_{\sU,i},\mX_{\sP,i},\mP_{T,i},\mP_{P,i}\}
  \end{align}
  \STATE Add \textbf{Pseudo Cell} $\hat v_i$ into $\hat\sV_M$
  \ENDFOR

  \STATE Set $\sV'_M=\sV_M-\bigcup_{i=1,\cdots,n}\sV_i+\hat\sV_M$ 
  \STATE Set $\sU'=\{u\in\sU|\exists v\in\sV'_M,(v,u)\in\sP\}$
  \STATE Initialize \textbf{Pseudo Pin} set: 
  \begin{equation}
    \hat\sP=\{(\hat v_i,u)|\exists v\in\sV_i,(v,u)\in\sP\wedge u\in\sU'\}
  \end{equation}
  \STATE Set $\sP'=\{(v,u)\in\sP|v\in\sV'_M\}\cup\hat\sP$
  \STATE Calculate $\mX'_\sV,\mX'_\sU,\mX'_\sP$ according to Appendix \ref{sec:feat-pseudo}
  \STATE Root graph $\gG_R$:
  \begin{equation}
      \gG_R=\{\sV_T\cup\sV_P\cup\sV'_M,\sU',\sP',\mX'_\sV,\mX'_\sU,\mX'_\sP,\mP_T,\mP_P\}
  \end{equation}
  \STATE {\bfseries Output:} sub-netlist graphs $\{\gG_i\}$ and coarsened netlist graph $\gG_R$
\end{algorithmic}
\end{breakablealgorithm}

\section{Cell-flow Construction}

The construction of cell-flow follows \cref{alg:cell-flow}.

\begin{breakablealgorithm}
  \caption{The algorithm of constructing cell-flow from a netlist graph.}
  \label{alg:cell-flow}
\begin{algorithmic}[1]
  \STATE {\bfseries Input:} netlist graph $\gG=\{\sV,\sU,\sP,\mX_\sV,\mX_\sU,\mX_\sP,\mP_T,\mP_P\}$
  \STATE Initialize flow edges $\sF=\emptyset$
  \STATE Initialize cell-incoming-edge function $f_{\rm INCOME}\in\sV\rightarrow\sF$
  \STATE Initialize edge-net function $f_{\rm NET}\in\sF\rightarrow\sU$
  \STATE Initialize $\hat\sV=\sV_T\cup\sV_P$ and $\hat\sU=\sU$
  \REPEAT
  \STATE Pop a cell $v$ from $\hat\sV$
  \FORALL{$u$ with $(v,u)\in\sP\wedge u\in\hat\sU$}
  \STATE Remove $u$ from $\hat\sU$
  \FORALL{$\overline v$ with $(\overline v,u)\in\sP\wedge\overline v\in\sV_M$}
  \STATE Add $\overline v$ to $\hat\sV$
  \STATE Add $(v,\overline v)$ to $\sF$
  \STATE Set $f_{\rm NET}((v,\overline v))=u$
  \IF{$f_{\rm INCOME}(\overline v)={\rm NULL}$}
  \STATE Set $f_{\rm INCOME}(\overline v)=(v,\overline v)$
  \ENDIF
  \ENDFOR
  \ENDFOR
  \UNTIL{$\hat\sV$ is empty}
  \STATE {\bfseries Output:} cell-flow $\sF$, function $f_{\rm INCOME}$ and $f_{\rm NET}$
\end{algorithmic}
\end{breakablealgorithm}

\section{Featurization of pseudo cells and pins}
\label{sec:feat-pseudo}

Although most of $\mX'_\sV,\mX'_\sU,\mX'_\sP$ can be inherited from original netlist graph $\gG$, we need to generate features for pseudo cells $\hat\sV_M$ and pseudo pins $\hat\sP$. Here, we regard a pseudo cell $\hat v_i$ as normal cell with width and height $(\sqrt{5S_i},\sqrt{5S_i})$, where $S_i=\sum_{v_j\in\sV_i}w_jh_j$ is the summation of cell areas inside $i$-th partition $\sV_i$. Pseudo pins are regarded as normal pins connected among nets and pseudo cells. Then we follow the cell/pin featurization in \cite{circuitgnn} to generate the features for pseudo cells and pins, and concatenate them to $\mX'_\sV,\mX'_\sP$.

\begin{table*}[htbp]
\caption{Runtime comparison between DREAMPlace and TransPlace on ISPD 2015}
\label{tab:app-detail-rt}
\centering
\begin{tabular}{lcccc}
\toprule
Netlist               & DREAMPlace & Ours(Inductive Placement) & Ours(Fine-tune) & Ours(Total) \\ \midrule
mgc\_des\_perf\_1     & 34.32      & 1.42                      & 35.83           & 37.25       \\
mgc\_des\_perf\_a     & 143.58     & 1.17                      & 70.08           & 71.25       \\
mgc\_des\_perf\_b     & 107.58     & 1.42                      & 88.59           & 90.01       \\
mgc\_edit\_dist\_a    & 65.93      & 1.52                      & 60.01           & 61.53       \\
mgc\_fft\_1           & 35.14      & 0.44                      & 10.61           & 11.05       \\
mgc\_fft\_2           & 41.37      & 0.35                      & 25.83           & 26.18       \\
mgc\_fft\_a           & 35.24      & 0.29                      & 11.35           & 11.64       \\
mgc\_matrix\_mult\_1  & 37.51      & 1.92                      & 20.04           & 21.96       \\
mgc\_matrix\_mult\_2  & 39.60      & 1.89                      & 18.65           & 20.54       \\
mgc\_matrix\_mult\_a  & 32.69      & 1.65                      & 16.89           & 18.54       \\
mgc\_pci\_bridge32\_a & 157.47     & 0.38                      & 155.79          & 156.17      \\
mgc\_pci\_bridge32\_b & 216.42     & 0.26                      & 147.31          & 147.57      \\
mgc\_superblue11\_a   & 97.39      & 11.40                     & 74.29           & 85.69       \\
mgc\_superblue12      & 70.33      & 15.43                     & 81.04           & 96.47       \\
mgc\_superblue14      & 35.33      & 7.11                      & 26.35           & 33.46       \\
mgc\_superblue16\_a   & 82.64      & 6.66                      & 109.08          & 115.74      \\
mgc\_superblue19      & 27.23      & 5.84                      & 24.97           & 30.81       \\
Average   Ratio       & 1.49       & 0.07                      & 0.93            & 1.00        \\ \hline
\end{tabular}
\end{table*}

\section{Optimality Demonstration}

\subsection{Optimality of Sub-netlist Graph Generation}
\label{sec:opt-hier}

The time complexity of \cref{alg:hier} is composed of three parts:
\begin{enumerate}
    \item Initialization of all $\sU_i$ and $\sP_i$ takes $O(|\sP|)$.
    \item Constructing sub-netlist graphs takes $\sum_i(|\sV_i|+|\sU_i|+|\sP_i|)=O(|\sV|+\eta|\sU|+|\sP|)$, with overlap ratio $\eta=\sum_i|\sU_i|/|\sU|$.
    \item Note that we set $f_{\rm BELONG}(\hat v_i)=0$, so we can identify $v\in\sV'_M$ and $v\in\sV_i$ in $O(1)$ time. $u\in\sU'$ can also be identified if we label the nets when calculating $\sU'$. Therefore, constructing root graph $\gG_R$ takes $O(|\sV|+|\sU|+|\sP|)$ in total.
\end{enumerate}

Time complexity of sub-netlist graph generation is $O(|\sV|+\eta|\sU|+|\sP|)$. With the help of placement prototype tool KaHyPar\cite{kahypar}, it can approximate $O(|\sV|+|\sU|+|\sP|)$ because the overlap among $\{\sU_i|i=1,\cdots,n\}$ is minor ($\eta\rightarrow1$). \cref{tab:opt-dataset} shows the overlap ratio $\eta$ in major datasets.

\begin{table}[htbp]
\caption{Overlap Ratio $\eta$ and Average Cell-path Length $\omega$ in \texttt{ISPD2015}.}
\label{tab:opt-dataset}
\centering
\begin{tabular}{lrrrr}
\hline
\multicolumn{1}{c}{Netlist} & \multicolumn{1}{c}{\# of cells} & \multicolumn{1}{c}{\# of nets} & \multicolumn{1}{c}{$\eta$} & \multicolumn{1}{c}{$\omega$} \\ \hline
mgc\_fft\_1 & 35K & 33K & 1.1262 & 3.1798 \\ 
mgc\_fft\_2 & 35K & 33K & 1.1257 & 3.1941 \\ 
mgc\_fft\_a & 34K & 32K & 1.0369 & 2.6055 \\ 
mgc\_fft\_b & 34K & 32K & 1.0399 & 2.5855 \\ 
mgc\_des\_perf\_1 & 113K & 113K & 1.1079 & 15.7122 \\ 
mgc\_matrix\_mult\_1 & 155K & 158K & 1.1177 & 5.4420 \\ 
mgc\_matrix\_mult\_2 & 155K & 158K & 1.1178 & 5.4057 \\ 
mgc\_matrix\_mult\_a & 149K & 154K & 1.0370 & 3.6518 \\ 
mgc\_superblue12 & 1293K & 1293K & 1.1509 & 20.8853 \\ 
mgc\_superblue14 & 634K & 620K & 1.1305 & 4.4139 \\ 
mgc\_superblue19 & 522K & 512K & 1.0969 & 4.4376 \\ \hline
\end{tabular}

\end{table}

\subsection{Optimality of Cell-flow Generation}
\label{sec:opt-flow}

\cref{alg:cell-flow} elaborates on how to generate cell-flow $\sF$ from a netlist graph $\gG$. Besides the initialization steps which cost $O(|\sV|+|\sU|)$, the bottleneck of its time consumption is the loop execution. For every $(v,u)\in\sP$ and net $u$ untraveled, we connect $v$ to almost every neighbor through $u$ and label $u$ as a traveled net. Because each net is traveled at most once, the loop body executes for no more than $\sum_{u\in\sU}|(v,u)\in\sP|=|\sP|$ times. Every statement in the loop body takes $O(1)$, so the time complexity of cell-flow generation is $O(|\sV|+|\sU|+|\sP|)$. Also note that the number of cell-flow edges equals to the execution times of loop body, i.e. $|\sF|=O(|\sP|)$.

\subsection{Optimality of SE(2)-invariant Encoding and Decoding}
\label{sec:opt-coding}

The time complexities of encoding and decoding are $O(|\sF|)=O(|\sP|)$ and $O(\omega|\sF|)=O(\omega|\sP|)$, respectively, where $\omega$ is the average length of cell-paths. The concrete values of $\omega$ in major datasets are listed in \cref{tab:opt-dataset}. For most datasets, $\omega$ is small, so it can be regarded as a constant.

\subsection{Time Complexity of Fine-tuning}
\label{sec:opt-ft}

In every epoch, the time complexities of calculating Wire-length, Density and Anchor objectives are $O(|\sP|)$, $O(|\sV|\log|\sV|)$ \cite{eplace}, and $O(|\sV|)$, respectively. So the whole fine-tuning has the complexity of $O(t_f(|\sP|+|\sV|\log|\sV|))$.

\section{Experimental Settings}

\begin{table*}[!htbp]
	\caption{Hyperparameters of circuit-adaptive fine-tuning on \texttt{ISPD2015}.}
	\label{tab:ours-ispd2015}
		\begin{tabular}{cccccccc}
			\toprule
			Netlist & Learning\_rate & Density\_weight & Iteration & RePlAce\_UPPER\_PCOF & Theta & Delta\_x & Delta\_y \\
			\midrule
			mgc\_des\_perf\_1     & 1e-1 & 8e-5 & 400   & 1.05 & 120 & 43.46   & 0       \\
			mgc\_des\_perf\_a     & 1e-2 & 8e-4 & 650   & 1.07 & 0   & 87.89   & 87.89   \\
			mgc\_des\_perf\_b     & 1e-2 & 8e-4 & 600   & 1.07 & 180 & -58.59  & 0       \\
			mgc\_edit\_dist\_a    & 1e-2 & 8e-4 & 650   & 1.07 & 120 & -78.12  & 78.12   \\
			mgc\_fft\_1           & 1e-2 & 8e-3 & 170   & 1.10 & 90  & 0       & -25.78  \\
			mgc\_fft\_2           & 3e-1 & 8e-3 & 150   & 1.10 & 90  & -33.40  & -33.40  \\
			mgc\_fft\_a           & 3e-1 & 8e-3 & 150   & 1.10 & 270 & -156.25 & 78.12   \\
			mgc\_matrix\_mult\_1  & 1e-1 & 8e-5 & 350   & 1.07 & 0   & 107.42  & -107.42 \\
			mgc\_matrix\_mult\_2  & 1e-2 & 8e-4 & 320   & 1.07 & 0   & -108.28 & 0       \\
			mgc\_matrix\_mult\_a  & 1e-1 & 8e-4 & 300   & 1.07 & 270 & -585.94 & 0       \\
			mgc\_pci\_bridge32\_a & 1e-2 & 8e-5 & 750   & 1.05 & 180 & 39.06   & -78.12  \\
			mgc\_pci\_bridge32\_b & 1e-2 & 8e-5 & 1000 & 1.05 & 90  & 78.12   & -156.25 \\
			mgc\_superblue11\_a   & 1e-2 & 8e-4 & 650   & 1.05 & 240 & -786.80 & 684.67  \\
			mgc\_superblue12      & 1e-2 & 8e-9 & 1000 & 1.06 & 0   & -276.80 & 201.62  \\
			mgc\_superblue14      & 1e-2 & 8e-4 & 350   & 1.06 & 0   & -253.05 & -368.09 \\
			mgc\_superblue16\_a   & 1e-2 & 8e-9 & 2000 & 1.06 & 0   & -144.81 & -157.15 \\
			mgc\_superblue19      & 1e-1 & 8e-5 & 400   & 1.05 & 270 & -199.38 & 403.24 \\
			\bottomrule
	\end{tabular}
\end{table*}
\begin{table*}[!htbp]
	\caption{Hyperparameters of circuit-adaptive fine-tuning on \texttt{DAC2012}.}
	\label{tab:ours-dac2012}
		\begin{tabular}{cccccccc}
			\toprule
			Netlist & Learning\_rate & Density\_weight & Iteration & RePlAce\_UPPER\_PCOF & Theta & Delta\_x & Delta\_y \\ \midrule
			superblue2  & 1e-2 & 3e-5 & 2000 & 1.05 & 270 & -240.16 & 0       \\
			superblue3  & 1e-2 & 8e-9 & 2000 & 1.06 & 120 & -498.95 & -323.44 \\
			superblue6  & 1e-2 & 8e-7 & 2000 & 1.05 & 240 & 0       & 0       \\
			superblue7  & 3e-4 & 1e-5 & 2000 & 1.05 & 270 & -155.33 & 0       \\
			superblue9  & 3e-3 & 3e-6 & 2000 & 1.05 & 270 & -132.71 & 0       \\
			superblue11 & 1e-2 & 1e-3 & 2000 & 1.05 & 90  & 0       & 0       \\
			superblue12 & 3e-2 & 8e-4 & 600  & 1.05 & 90  & -138.40 & 0       \\
			superblue14 & 1e-3 & 3e-5 & 1000 & 1.05 & 90  & 0       & 0       \\
			superblue16 & 4e-2 & 6e-5 & 1250 & 1.05 & 270 & 0       & -157.15 \\
			superblue19 & 1e-2 & 8e-5 & 2000 & 1.05 & 270 & -99.69  & 0       \\ \bottomrule
	\end{tabular}
\end{table*}

\subsection{\gnn{} Training}\label{sec:app-gnn-training}


\gnn{} is trained to imitate preplaced circuits generated by DREAMPlace \cite{dreamplace}. These circuits consist of netlist graphs $\gG$ with absolute positions for movable cells $\hat\mP_M$. We encode these absolute positions into ground-truth relative positions, which serve as training labels.

The hidden layer dimensions of \textit{cell}, \textit{net}, and \textit{pin} are set to $D_\sV=64$, $D_\sU=64$, and $D_{\sP}=8$, respectively. We set message passing layers $L=3$ and loss weights $\lambda_\theta=1.0,\lambda_{\rm D}=8e-6,\lambda_{\rm A}=1e-2$.

When optimizing \textbf{\gnn} with \texttt{Adam Optimizer}, we use learning rate $\Lambda_t=5e-5$, learning rate decay $\Delta\Lambda_t=1e-3$, weight decay $\eta_t=5e-4$, and training epoch $\tau_t=500$. The training loss is determined by the difference between \gnn{} outputs and ground truth relative positions:
\begin{equation}
	\gL_{\rm train} = {\rm Smooth\text{-}L1}(\bm{\rho},\bm{\hat\rho})+\lambda_\theta\cdot{\rm Smooth\text{-}L1}(\Delta\vtheta,\Delta\hat\vtheta),
	\label{eq:gnn-train}
\end{equation}
where ${\rm Smooth\text{-}L1}$ is a robust ${\rm L1}$ loss function, and $\lambda_\theta$ weighs two loss terms, which is set to 0.1. Our source code is available at \href{https://github.com/sorarain/TransPlace}{https://github.com/sorarain/TransPlace}.

\subsection{Optimization Settings}

We use default baseline settings. Both DREAMPlace and circuit-adaptive fine-tuning use \texttt{NAG Optimizer} \cite{eplace} for a fair comparison, adopting the $\gamma$ and $\lambda_{\rm D}$ adjustment strategy from Lu et al. \cite{eplace}. Routability optimization via cell inflation follows Lin et al. \cite{dreamplace2.0} with default hyperparameters. Timing optimization differs from DREAMPlace \cite{dreamplace4}, evaluating timing metrics and updating net weights every 15 iterations after placement overflow reaches $st_{overflow}=0.5$ or $st_{iteration}=500$. For \texttt{DAC2012}, \texttt{ISPD2015}, and \texttt{ISPD2019}, the rudy map guides cell inflation. Our fine-tuning parameters are given in \cref{tab:ours-ispd2015}, \cref{tab:ours-dac2012}, \cref{tab:ours-ispd2019}, and \cref{tab:iccad2015}; ``Iteration'' is the maximum iteration limit, a variable setting up an early stop strategy \cite{eplace, dreamplace}.

\begin{table*}[!htbp]
	\caption{Hyperparameters of circuit-adaptive fine-tuning on \texttt{ISPD2019}.}
	\label{tab:ours-ispd2019}
		\begin{tabular}{ccccccc}
			\toprule
			Netlist        & Learning\_rate & Density\_weight & Iteration & Theta & Delta\_x & Delta\_y \\ \midrule
			ispd19\_test1  & 2e-3           & 5e-9            & 550       & 90    & 0        & -28.13   \\
			ispd19\_test2  & 3e-4           & 1e-7            & 600       & 90    & 0        & -113.91  \\
			ispd19\_test3  & 2e-3           & 5e-9            & 550       & 180   & -27.01   & 0        \\
			ispd19\_test4  & 3e-4           & 1e-3            & 550       & 180   & -156.25  & -151.37  \\
			ispd19\_test6  & 1e-2           & 8e-5            & 1000      & 0     & -173.83  & -149.53  \\
			ispd19\_test7  & 1e-2           & 8e-5            & 1000      & 0     & -217.29  & -186.80  \\
			ispd19\_test8  & 1e-2           & 8e-5            & 1000      & 0     & -344.73  & -224.06  \\
			ispd19\_test9  & 1e-2           & 1e-7            & 700       & 0     & -384.38  & -300.23  \\
			ispd19\_test10 & 1e-2           & 1e-5            & 550       & 180   & 0        & 0        \\ \bottomrule
	\end{tabular}
\end{table*}

\begin{table*}[!htbp]
	\caption{Hyperparameters of circuit-adaptive fine-tuning on \texttt{ICCAD2015}.}
	\label{tab:ours-iccad2015}
		\begin{tabular}{cccccccc}
			\toprule
			Netlist & Learning\_rate & Density\_weight & Iteration & RePlAce\_UPPER\_PCOF & Theta & Delta\_x & Delta\_y \\ \midrule
			superblue1  & 1e-02 & 8e-03 & 1,000 & 1.06 & 72  & -219.38 & 80.38  \\
			superblue3  & 1e-02 & 8e-05 & 1,000 & 1.05 & 0   & 0       & 0      \\
			superblue4  & 1e-02 & 8e-05 & 1,000 & 1.05 & 300 & 581.99  & 0      \\
			superblue5  & 1e-02 & 8e-05 & 1,000 & 1.05 & 0   & 0       & 0      \\
			superblue7  & 1e-02 & 8e-05 & 1,000 & 1.06 & 144 & -310.66 & 278    \\
			superblue10 & 1e-02 & 8e-05 & 1,000 & 1.05 & 120 & 424.84  & 302.08 \\
			superblue16 & 1e-02 & 8e-05 & 1,000 & 1.05 & 144 & -144.81 & 78.57  \\
			superblue18 & 1e-02 & 8e-05 & 1,000 & 1.05 & 0   & 0       & 0     \\ \bottomrule
	\end{tabular}
\end{table*}
\section{Detailed Runtime Analysis}
\label{sec:app-rt-anla}
Here we detailed analysis of the runtime of baselines and TransPlace. We compare DREAMPlace with our method on ISPD2015 in Tab \ref{tab:app-detail-rt}.

\section{Ablation Study}
\label{sec:app-abla}
\begin{table*}[h]
        \caption{Performance comparison of TransPlace, TransPlace without fine-tuning, and DREAMPlace (i.e. fine-tuning with random initialization) on ISPD 2015.}
        \label{tab:app-abla}
	\centering
	\resizebox{\textwidth}{!}{
	\begin{tabular}{ccccccccc}
\hline
\multicolumn{1}{c}{\multirow{2}{*}{Netlist}} &
  \multicolumn{1}{c}{\multirow{2}{*}{\# of cells}} &
  \multicolumn{1}{c}{\multirow{2}{*}{\# of nets}} &
  \multicolumn{3}{c}{OVFL$\downarrow$} &
  \multicolumn{3}{c}{RWL($\times10^6um$)$\downarrow$} \\ \cline{4-9} 
\multicolumn{1}{c}{} &
  \multicolumn{1}{c}{} &
  \multicolumn{1}{c}{} &
  Ours &
  DREAMPlace &
  Ours(w.o. fine-tune) &
  Ours &
  DREAMPlace &
  Ours(w.o. fine-tune) \\ \hline
mgc\_des\_perf\_1     & 113K  & 113K  & 24    & 41    & 1468276  & 1.81  & 1.70  & 9.24  \\
mgc\_des\_perf\_a     & 109K  & 110K  & 11841 & 13123 & 1155489  & 3.22  & 3.06  & 9.83  \\
mgc\_des\_perf\_b     & 113K  & 113K  & 7     & 13    & 1822843  & 2.59  & 2.38  & 13.93 \\
mgc\_edit\_dist\_a    & 130K  & 131K  & 13738 & 16251 & 1822843  & 6.15  & 6.18  & 13.93 \\
mgc\_fft\_1           & 35K   & 33K   & 5     & 19    & 5844     & 0.70  & 0.65  & 1.41  \\
mgc\_fft\_2           & 35K   & 33K   & 2     & 5     & 141      & 0.72  & 0.65  & 1.05  \\
mgc\_fft\_a           & 34K   & 32K   & 2604  & 3244  & 171367   & 1.01  & 0.99  & 2.98  \\
mgc\_matrix\_mult\_1  & 155K  & 158K  & 5     & 9     & 330279   & 3.03  & 3.01  & 7.61  \\
mgc\_matrix\_mult\_2  & 155K  & 158K  & 0     & 12    & 220378   & 2.98  & 3.00  & 7.00  \\
mgc\_matrix\_mult\_a  & 149K  & 154K  & 4477  & 6802  & 1107320  & 3.91  & 4.02  & 12.42 \\
mgc\_pci\_bridge32\_a & 30K   & 30K   & 2776  & 3376  & 95997    & 0.77  & 0.77  & 2.25  \\
mgc\_pci\_bridge32\_b & 29K   & 29K   & 2419  & 2921  & 150625   & 1.01  & 1.01  & 2.30  \\
mgc\_superblue11\_a   & 954K  & 936K  & 561   & 899   & -        & 44.46 & 44.04 & -     \\
mgc\_superblue12      & 1293K & 1293K & 67690 & 99641 & -        & 40.98 & 40.86 & -     \\
mgc\_superblue14      & 634K  & 620K  & 1283  & 1690  & -        & 30.05 & 29.54 & -     \\
mgc\_superblue16\_a   & 698K  & 697K  & 26894 & 27862 & -        & 33.59 & 34.01 & -     \\
mgc\_superblue19      & 522K  & 512K  & 110   & 432   & 5311769  & 20.76 & 20.47 & 52.63 \\
Average   Ratio       &       &       & 1.00  & 2.03  & 25753.73 & 1.00  & 0.97  & 2.47  \\ \hline
\end{tabular}
}
\end{table*}

Here we present the ablation study of TransPlace in Tab \ref{tab:app-abla}, where "-" denotes the placement results fail to route. From the result, we can infer that fine-tuning is crucial to guarantee reasonable placement, and inductive placement can further improve placement quality.

\end{document}